\documentclass[
]{style/ceurart}

\usepackage{listings}
\usepackage{graphicx}
\usepackage{booktabs}

\begin{document}

\copyrightyear{2025}
\copyrightclause{Copyright for this paper by its authors.
  Use permitted under Creative Commons License Attribution 4.0
  International (CC BY 4.0).}

\conference{CLEF 2025 Working Notes, 9 -- 12 September 2025, Madrid, Spain}

\title{Pun Intended: Multi-Agent Translation of Wordplay with Contrastive Learning and Phonetic-Semantic Embeddings for CLEF JOKER 2025 Task 2}

\author[1]{Russell Taylor}[
    orcid=0009-0007-0702-2375,
    email=rdtaylorjr@gatech.edu,
]
\cormark[1]

\author[1]{Benjamin Herbert}[%
orcid=0009-0009-0179-3835,
email=bherbert6@gatech.edu,
]

\author[1]{Michael Sana}[%
orcid=0009-0007-6608-7347,
email=msana3@gatech.edu,
]

\address[1]{Georgia Institute of Technology, North Ave NW, Atlanta, GA 30332}
\cortext[1]{Corresponding author.}

\begin{abstract}
Translating wordplay across languages presents unique challenges that have long confounded both professional human translators and machine translation systems. This research proposes a novel approach for translating puns from English to French by combining state-of-the-art large language models with specialized techniques for wordplay generation. 

Our methodology employs a three-stage approach. First, we establish a baseline using multiple frontier large language models with feedback based on a new contrastive learning dataset. Second, we implement a guided chain-of-thought pipeline with combined phonetic-semantic embeddings. Third, we implement a multi-agent generator-discriminator framework for evaluating and regenerating puns with feedback.

Moving beyond the limitations of literal translation, our methodology's primary objective is to capture the linguistic creativity and humor of the source text wordplay, rather than simply duplicating its vocabulary. Our best runs earned first and second place in the CLEF JOKER 2025 Task 2 competition where they were evaluated manually by expert native French speakers.

This research addresses a gap between translation studies and computational linguistics by implementing linguistically-informed techniques for wordplay translation, advancing our understanding of how language models can be leveraged to handle the complex interplay between semantic ambiguity, phonetic similarity, and the implicit cultural and linguistic awareness needed for successful humor.
\end{abstract}


\maketitle

\section{Introduction}

Training language models to translate puns is difficult for several reasons: 

First, most language models are designed to identify linguistic and semantic patterns and probabilistically eliminate outliers \cite{baziotis2023automatic}. But puns rely on both semantic ambiguities and linguistic discontinuities in order to produce humor. It is precisely the presence of some linguistic incongruity that makes a pun surprising or clever \cite{aarons2017puns}. And without it, computational approaches to humor often fail. 

Second, neural machine translation models have advanced rapidly in recent years and have achieved impressive results on machine translation tasks. But these models are traditionally trained using loss functions and evaluation metrics that reward direct, literal translations. There often does not exist a homonym in a target language that has precisely the same meanings as a homonym in a source language. So, a machine translation model may effortlessly choose a gloss, and inadvertently destroy the wordplay.

Third, generating a pun in a new language requires broad awareness of both the linguistic and cultural nuances of the target milieu. Producing original humor in any context is a difficult task, even for humans. Most jokes that people tell are simply repeated versions of jokes they have heard. Remarkably few talented or well-trained humans are able to produce truly original humor \cite{miller2019punsters}. 

Fourth, translating humor has long been recognized by professional translators as a task so difficult, that it is often dismissed as impossible \cite{low2011translating}. 

However, with the rapid advancement of frontier large language models (LLMs), which are trained on datasets that contain enormous amounts of cultural and linguistic data, including the semantic and linguistic incongruities of humor, the potential for quality machine-translated puns is greater than ever before. This paper proposes to take advantage of both the latest advances in LLMs and the latest state-of-the-art approaches for single-language pun generation, and apply them to the task of translating puns from English into French. 

This paper will address the following research questions:

\begin{enumerate}
\item How well do the latest large language models, supplemented by a discriminator model trained using contrastive learning, compare to the previous state-of-the-art for translating puns from English to French?
\item How well does a guided chain-of-thought pipeline with trained phonetic-semantic embeddings compare to the previous state-of-the-art for translating puns from English to French?
\item How well does a multi-agent generator-discriminator pipeline with feedback compare to the previous state-of-the-art for translating puns from English to French?
\end{enumerate}
\section{Related Work}

The literature relevant to this task includes research on linguistic approaches to humor, approaches to professional human translation of wordplay, machine translation of wordplay, wordplay generation, and evaluation of generated wordplay.

\subsection{Linguistics of Humor}

The linguistics of humor is a rich field at the intersection of pragmatics, semantics, sociolinguistics, and cognitive linguistics. Several major theories attempt to explain how humor operates in language. 

The General Theory of Verbal Humor (GTVH), proposed by Attardo and Raskin, aims to describe the linguistic functions that result in humor. Central to the theory is the concept of script opposition, where two incompatible frames of reference are juxtaposed. This incongruous juxtaposition creates surprise and therefore humor \cite{attardo1991script}. Veisbergs argues that GTVH allows for identifying which joke elements are essential, and shows how shifts in the language or logical mechanism can still preserve humor as long as script opposition is retained \cite{veisbergs1997contextual}.

The primary alternative approach to GTVH in the literature is Relevance Theory. While GTVH is rooted in structuralist linguistics, Relevance Theory is based in pragmatics and cognitive linguistic approaches. It suggests that humor involves violations of relevance expectations, often leading to reinterpretation or cognitive reprocessing. In support of this approach, Yus specifically argues that pun translation is successful when it results in a similar level of surprise and reinterpretation \cite{yus2003humor}.

Aarons argues that successful puns require both ambiguity (script overlap) and incongruity (script oppositeness). But while GTVH can be useful for describing humor linguistically, it does not imply that users of humor are aware of the linguistic features at play. Aarons argues that tacit linguistic knowledge is required by both the speaker and hearers of puns in order for the pun to succeed \cite{aarons2017puns}. Based on this insight, we believe it will be important to make use of the tacit linguistic knowledge of LLMs, as opposed to more specialized neural machine translation models.

\subsection{Professional Human Translation of Wordplay} 

The most widely cited framework for translating puns, proposed by Delabastita in 1996, identifies eight strategies that translators might employ \cite{delabastita1996introduction}. A volume of essays published the following year, entitled \textit{Traductio: Essays on Punning and Translation}, explores these approaches from many different perspectives and domains \cite{delabastita1997introduction}. While this work has been influential, it does not provide a step-by-step process for translating puns.

Low fills this gap with an insightful essay that argues persuasively against the defeatist attitude present among many translators that pun translation is often impossible \cite{low2011translating}. He proposes a systematic methodology for translating puns that he represents through the visual of polygons.

He begins with what he describes as a "square" translation of the pun, where the first and second meanings of the homonym are simply translated directly, and the four corners of the square represent the two source language meanings and two target language meanings. However, in many cases, no homonym exists in the target language with similar semantic ranges to the two original meanings. In such cases, Low employs a "pentagon" translation, where he finds an alternative word in the target language that is similar phonetically to one of the directly translated words, and also similar semantically to the other directly translated word. This added step is represented by the fifth vertex of the pentagon. When a suitable homonym is still not found, the same search can be performed for in the opposite direction. This becomes the sixth vertex of a "hexagon". These steps may be iterated as long as the translator has the will and patience to do so. Low argues that this approach leads to a far higher likelihood of finding a suitable pun in the target language that closely matches the intent of the original.

Because Low so carefully and systematically described this approach, it is not hard to imagine a computational implementation of this algorithm.

\subsection{Machine Translation of Wordplay}

The JOKER lab at the Conference and Labs of the Evaluation Forum (CLEF) has been the primary venue for research on machine translation of wordplay since 2022 \cite{ermakova2025overview}. The best result based on human evaluation in 2023 was only 6\% of generated translations containing wordplay and preserving the meaning of the source puns over the total test set \cite{ermakova2023overview}. This relatively poor showing highlights the extent to which language models still struggle to translate idiomatic language as well as the significant room for improvement in this area.

In a position paper, Miller proposes a theoretically possible computational approach to the translation of puns: 1. scan the source text and flag possible puns, 2. identify the incongruous meanings using word sense and semantic role knowledge-bases, 3. look up translations of the pun’s two meanings and search for closely related senses in the target language, 4. search among those results to find phonetically similar candidates, 5. repeat the above to generated a set of candidate translations, 6. rank the candidate translations and select the most promising \cite{miller2019punsters}. 

Miller's proposal was written in 2019, when transformers were still in their infancy and very large language models had yet to make their debut. At the time, Miller's approach was perhaps not yet feasible, but with the latest LLMs, we hypothesize that each of these steps can be achieved with a high degree of accuracy.

\subsection{Wordplay Generation}

Significant advances have been made in recent years in the area of English-only pun generation. Several insights are directly applicable to the pun translation task.

Xu, et al. use chain-of-thought prompting to evaluate several LLMs on pun recognition, explanation, and generation tasks. For pun recognition, they prompt multiple recent LLMs to identify whether a given text is a pun or non-pun. They find significant variance with different prompts and emphasize the importance of experimentation to find the best prompt for the use case. For pun explanation, they ask the LLMs to identify the pun word and its alternative meaning. They find that LLMs can accurately recognize pun words in sentences, but struggle to identify the alternative meanings, and also that LLMs are worse at explaining heterographic puns than homographic puns. For pun generation, they provide a homographic or heterographic word pair, and ask the LLMs to generate a pun sentence, and find that LLMs are better and generating homographic puns and tend to include both words when prompted to generate heterographic puns  \cite{xu2024a}.

Both Zhong, et al.\cite{zhong2024lets} and Wang, et al.\cite{wang2024innovative} aim to improve upon chain-of-thought prompting for humor generation with what they call leap-of-thought prompting. The idea is to force the LLM to use randomized inputs in its generation in order to inspire "creative" and out-of-the-box thinking. Wang, et al. extend this approach further using a multi-agent GAN-inspired setups for both dataset creation, and pun generation. During the pun generation stage, a generator model generates two independent solutions, a pre-trained evaluator chooses between them, then a third model provides a rationale for the choice. They report state-of-the-art results on the single-language pun generation task.

Zeng, et al. model pun sentences using semantic trees and pruning, then create a contrastive learning dataset in order to discriminate between puns and non-puns. Finally, they generate puns using a GAN where the generator uses the semantic trees technique, and the discriminator is trained on the contrastive learning dataset \cite{zeng2024barking}.

Several other approaches to pun generation have been influential in recent years. Sun, et al. augment the SemEval 2017 Task 7 dataset with human annotations about each pun to produce the ExPUNations dataset used by multiple papers already cited \cite{sun2022expunations}. Mittal, et al. experiment with lookups for co-occurring words in natural language datasets for appropriate context words relevant to each of the two meanings of a homonym. They use those context words to generate suitable pun sentences \cite{mittal2022ambipun}. He, et al. build a framework based on an observed pattern that a pun word's literal meaning is often supported by context words in the distant context of the sentence, while the pun word's alternative meaning is often supported by an idiomatic phrase in its immediate context \cite{he2019pun}. Tian, et al. aim to combine the approaches of Mittal, et al. and He, et al. into a unified framework suitable for both homographic and homophonic puns \cite{tian2022unified}.

Finally, Sharma, et al. design a method for creating phonetic embeddings using freely available IPA datasets for both Hindi and English. They test their embeddings using the Joker English dataset \cite{sharma2021phonetic}. We use their methodology to create French phonetic embeddings for use in our pun translation pipeline.

\subsection{Evaluation of Generated Wordplay}

Evaluation of generated and translated wordplay is particularly challenging. Traditional metrics like BLEU and BERTScore reward direct, literal translations, and therefore penalize outputs that include idiomatic language.

Wang et al. aim to address this problem by developing a framework for evaluating metaphor translations across languages. They create a corpus called MMTE, which highlights four critical evaluation criteria: quality, metaphorical equivalence, emotion, and authenticity. They find that LLM evaluations using these criteria can produce results comparable to human evaluation \cite{wang2024mmte}.

Baziotis et al. propose an algorithm for identifying literal translation errors in idioms. Their Literal Translation Error Rate (LiTER) methodology systematically identifies when figurative expressions are erroneously translated word-for-word \cite{baziotis2023automatic}. This methodology can be used where pun location data is present, so it should be useful for this proposed research.

Góes et al. offer an innovative approach to automated joke evaluation through their "Crowd Score" method. Using multiple LLMs with different "personalities" based on humor types (affiliative, self-enhancing, aggressive, and self-defeating), they create a diverse panel of LLM evaluators that collectively assess joke quality. Their approach accounts for the subjective nature of humor and can be adapted to account for culture-specific sensibilities \cite{goes2022crowd}.

\section{Methodology}

First we describe the data and resources used and then we describe the procedure we followed. Our procedure is divided into three parts which correspond to our three research questions: 1. Baseline pun generation with contrastive learning, 2. Guided chain-of-thought with phonetic-semantic embeddings, and 3. Multi-agent evaluation of generated puns.

\subsection{Data and Resources}

In addition to our primary dataset, we used data from multiple other sources, including existing datasets, generated data, and manual annotations. Here we describe the datasets and resources that we used in this research project.

\subsubsection{The 2025 JOKER Task 2 Wordplay Translation dataset}
The CLEF JOKER 2025 Task 2: Wordplay Translation dataset was our primary dataset \cite{ermakova2025overview}\cite{ermakova2025overview-task2}. The training data consists of 1,405 curated English pun sentences and 5,838 French translations of those pun sentences. The number of French translations per English pun ranges from 1 to 29. The test data consists of 376 English pun sentences and 832 corresponding French translations of those pun sentences. However, the organizers of the JOKER shared task do not publicly release the French translations for the test data. The precise identities of the 376 English pun sentences are obfuscated by placing them within a set of 4,537 English pun sentences, which is released to shared task participants.

\subsubsection{The 2023 JOKER Task 2 Pun Location and Interpretation dataset}
We also used the CLEF JOKER 2023 Task 2 Pun Location and Interpretation dataset \cite{ermakova2023overview}. This data identifies the pun word in each pun sentence and also includes a definition for each of the two meanings of the pun word. The training data contains entries for 2,315 English pun sentences and 2,000 French pun sentences, and the test data contains entries for 1,205 English pun sentences and 4,655 French pun sentences. There is not a 100\% correspondence between these pun sentences and the sentences in our primary dataset, but there was enough overlap that these annotations saved us time in our manual annotation of our primary data.

\subsubsection{Manually annotated pun identifications and types}
We created a new set of manual annotations for each of the 1405 English pun sentences in our primary dataset. These annotations include the pun word, the pun type (homographic or homophonic), the implied homophone (if applicable), each of the two meanings of the pun word and/or its implied homophone, and any supporting context words for each of the two meanings. We used this dataset to evaluate the performance of multiple models on the preliminary pun location and identification task.

\subsubsection{Generated French non-pun examples for contrastive learning}
We also created a new contrastive learning dataset based on the 5,838 French translations in our primary dataset. To create this, we prompted \texttt{gemini-2.5-flash-preview-05-20} to replace the pun word in each sentence with a different word so that the sentence no longer contains any sort of wordplay. Using a second prompt, we verified that the generated sentence is not a pun, and retried until this test passed. We then combined the pun and non-pun sentences, added binary target information, and trained a model to correctly distinguish between puns and non-puns. We took inspiration from Ermakova et al.\cite{ermakova2023joker} for  “destroying” puns and Zeng et al.\cite{zeng2024barking} for contrastive learning.

\subsubsection{Semantic and phonetic embeddings datasets}
We used the pre-trained \texttt{cc.fr.300.bin} French semantic embeddings from FastText \cite{grave2018learning}. In order to train phonetic embeddings, we used the Lexique database, which contains lemma, number of syllables, grammatical category, phonological representation for 140,000 French words. We also used PanPhon, a database relating over 5,000 IPA segments to 21 subsegmental articulatory features \cite{panphon2016}.

\subsubsection{Resources}

We used LangChain to access multiple large language models via their official APIs. These models include: 

\begin{itemize}
\item OpenAI's \texttt{o3}, \texttt{o4-mini-2025-04-16}, and \texttt{gpt-4.1}
\item Google's \texttt{gemini-2.5-pro-preview-05-06} and \texttt{gemini-2.5-flash-preview-05-20}
\item Anthropic's \texttt{claude-sonnet-4-20250514}
\item Mistral's \texttt{mistral-medium-2505}
\item DeepSeek's \texttt{deepseek-reasoner}
\end{itemize}
We chose these models based on benchmark results reported at \url{https://artificialanalysis.ai} at the time of our experiments in May/June 2025. 

We also experimented with Google's \texttt{google-translate} API for translating pun words and meanings. For evaluating translation of pun words and meanings we used SentenceTransformers with the French/English \texttt{Lajavaness/bilingual-embedding-large} model \cite{conneau2019unsupervised}.

\subsection{Part 1: Baseline Pun Generation with Contrastive Learning}

\begin{figure}[ht]
    \centerline{\includegraphics[width=360pt]{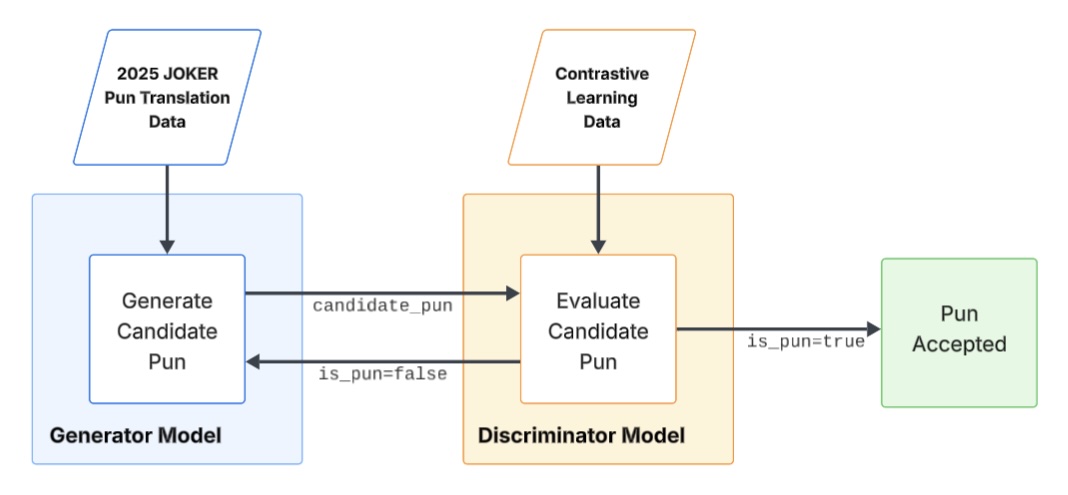}}
    \caption{Baseline pun cross-lingual pun generation with contrastive learning design.}
    \label{fig-contrastive}
\end{figure}

\subsubsection{Exploratory data analysis and data cleaning}

We cleaned the data by correcting erroneous and extraneous punctuation, removing hash tags, lower-casing words that were arbitrarily capitalized, and replacing proper names with pronouns because in early experiments LLMs gave too much weight to the proper names. We identified these categories based on a manual review of the dataset.

\subsubsection{Unsupervised cross-lingual LLM pun generation}
Our baseline solution to the pun translation task was to use a simple prompt  with multiple LLMs. We specifically chose LLM prompting, rather than fine-tuned neural machine translation (NMT) models, because the latter are explicitly trained using loss functions that reward literal translations and tend to eliminate outliers, while puns rely on non-literal use of language and linguistic non-conformity. Compared to machine translation models, LLMs excel at introducing creativity and spontaneity into their outputs due to the vast amounts of humorous text included in their training data.

We anticipate that this approach will produce worse results when evaluated using standard automated metrics like BLEU and BERTScore, since those metrics similarly favor literal translations over non-literal ones.

Our prompts included three parts: First, we specifically avoided using the word “translate” when prompting the models. We found in many early experiments that asking LLMs to translate caused them to default to the typical behavior of translating literally at the cost of preserving the wordplay. Second, following Mittal et al.\cite{mittal2022ambipun}, we prompted the models to choose a homonym where the first meaning is related to the broader context and the second meaning is part of an idiomatic phrase. Third, we instructed the model to produce a pun where both meanings are obvious and funny to a native French speaker. Without this, the LLMs tended to produce outputs where one of the meaning was so obscure as to be unnoticeable. Our code and full prompts are publicly available at \url{https://github.com/dsgt-arc/joker-2025}.

\subsubsection{Contrastive learning}

We used a second LLM as a discriminator model and asked it to identify whether each generated sentence contained a pun (1) or not (0). If its response was 0, we prompted the generator model again, and repeated up to 10 times. After 10 retries, we accepted the generated sentence with the annotation \texttt{is\_pun=0}. We chose \texttt{gemini-2.5-flash-preview-05-20} for this purpose because of its speed and in order to avoid bias since it was not one of our generator models.

To improve the performance of the discriminator model we used contrastive learning, following Zeng et al\cite{zeng2024barking}. We created a new contrastive learning dataset containing 5,838 positive examples of French puns and 5,838 negative examples of French puns, as detailed above. We then randomly selected 25 positive examples and 25 negative examples and trained our discriminator model using several-shot prompting.

\subsection{Part 2: Guided Chain-of-Thought with Phonetic-Semantic Embeddings}

\begin{figure}[ht]
    \centerline{\includegraphics[width=430pt]{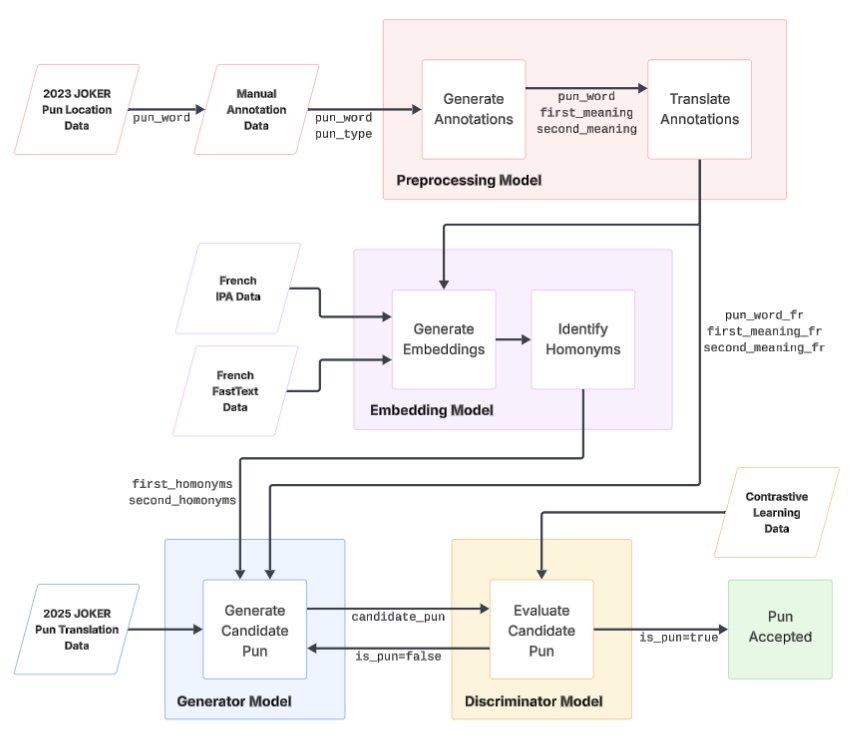}}
    \caption{Guided chain-of-thought with phonetic-semantic embeddings design.}
    \label{fig-cot_embedding}
\end{figure}

\subsubsection{Identification of pun word and pun type}

Using the cleaned English puns from the previous step, we prompted LLMs to identify the pun word (homonym) in each English pun sentence, as well as the pun type (homographic or homophonic).

To evaluate the performance of the LLMs on these tasks, we manually annotated the pun word and pun type for each of the 1405 English puns in the JOKER training dataset. For pun word, we started with the CLEF 2023 JOKER Task 2 Pun Location and Interpretation dataset \cite{ermakova2023overview}, made corrections as needed, and manually identified the pun word in 289 cases where data was not already present. Then we calculated accuracy, precision, recall, and F1-scores for the generated versus manual annotations.

\subsubsection{Translation of pun word and meanings}
As part of the same prompt for identifying the pun word and pun type, we also asked the LLMs to generate a list of synonyms for each of the two meanings of the pun. Next, we prompted multiple LLMs to translate the identified pun word and each element in the two lists of synonyms into French. 

To evaluate the performance of the LLMs, we embedded each word using SentenceTransformers with the French/English \texttt{Lajavaness/bilingual-embedding-large} model \cite{conneau2019unsupervised}. Then we calculated the cosine similarity for each word pair and took the mean for all translated words for each model. Additionally, we calculated the variance, top and bottom quartiles for cosine similarity, and percentage of words that were left untranslated for each model.

Additionally, we prompted \texttt{o4-mini} to identify whether the translated French pun word is a homonym, whether its meaning overlaps with the meanings of the words in the first translated synonyms list, and whether its meaning overlaps with the meanings of the words in the second synonyms list.

\subsubsection{Phonetic-semantic embeddings}

We follow the methods outlined in Sharma et al. to generate a learned continuous embedding space for the French language \cite{sharma2021phonetic}. They proposed a method for calculating the phonetic similarity of words that accounts for human perception of sounds. They used these similarity scores to construct a continuous embedding space for use in downstream phonology tasks. While their work demonstrated the effectiveness of this approach for English and Hindi, we apply their method to French.

The Lexique database contains 140,000 words of the French language, along with various information such as associated lemma, the number of syllables, the grammatical category, and phonological representation. 

PanPhon is a database relating over 5,000 IPA segments to 21 subsegmental articulatory features \cite{panphon2016}. Using PanPhon, we converted the French words and associated phonological IPA representation contained in the Lexique database into sequences of bigrams. The phonetic similarity of two feature sets of bigrams F(Pa) and F(Pb) can be computed using Jaccard similarity:

\[
S((P_{a1}, P_{a2}), (P_{b1}, P_{b2})) =
\frac{\left| F(P_{a1}, P_{a2}) \cap F(P_{b1}, P_{b2}) \right|}
     {\left| F(P_{a1}, P_{a2}) \cup F(P_{b1}, P_{b2}) \right|}
\tag{1}
\]

The method accounts for both phoneme similarity and the order in which phonemes appear by aligning bigram sequences using a dynamic programming algorithm. This algorithm evaluates all possible ways to align the bigrams of two words and selects the alignment that produces the highest cumulative similarity score. At each step, it can match two bigrams if they are similar, skip a bigram from one word, or skip from the other. Once the optimal alignment is found, the total score is normalized by the length of the longer sequence. This results in a final similarity score between 0 and 1, making it possible to compare scores across words of different lengths. 

We then train a BiLSTM encoder to learn a phonetic embedding space, using the similarity scores as supervision. We used an output layer of size 300 to match the dimensionality of the FastText semantic embeddings.

After creating the phonetic embeddings, we concatentated them with the pre-trained \texttt{cc.fr.300.bin} French semantic embeddings from FastText \cite{grave2018learning}. This created a combined phonetic-semantic vector space in which we could search for words that were semantically similar to one input word and phonetically similar to another input word.

For inference, we embedded the pun word and each word in both lists of meanings in the phonetic vector space, and we embedded the words in each list of meanings in the semantic vector space and took the mean for each meaning. We then searched in both directions: 1. combining the semantic embedding from the first meaning with the phonetic embeddings for each of the second meaning words, and 2. combining the semantic embedding from the second meaning with the phonetic embeddings for each of the first meaning words. 

We took the top k=2 results for each pair of semantic and phonetic embeddings with the constraint that the cosine similarity between the found word and both of the inputs must be > 0.75. This may be represented by:

\[
\frac{\vec{w}_{sem} \cdot \vec{S}}{\|\vec{w}_{sem}\| \|\vec{S}\|} > 0.75 \quad \text{and} \quad \frac{\vec{w}_{phon} \cdot \vec{P}}{\|\vec{w}_{phon}\| \|\vec{P}\|} > 0.75
\tag{2}
\]

\noindent where $\vec{w}$ is the combined 600-dimensional embedding of a candidate word, $\vec{w}_{sem}$ is the first 300 dimensions of $\vec{w}$ representing the word's semantic embedding, $\vec{w}_{phon}$ is the final 300 dimensions of $\vec{w}$ representing the word's phonetic embedding, $\vec{S}$ is the 300-dimensional input semantic vector, and $\vec{P}$ is the 300-dimensional input phonetic vector.

\subsubsection{Guided chain-of-thought}

To generate the puns, we used three distinct prompts depending on the pun type and the semantic overlap between the pun word translated to French and each of the two lists of synonyms translated to French. 

If the pun type is homographic, if the translated pun word is a homonym and if its meaning overlaps with both of the meanings represented by the translated lists of synonyms, we instructed the model to use that pun word in its generated pun sentence. 

If the pun type is homographic, but the other criteria are not met, we instructed the model to find a French homonym with two meanings similar to the meanings represented by the two lists of translated synonyms.

If the pun type is homophonic, we instructed the model to find two words that sound alike where each has a meaning similar to one of the lists of translated synonyms.

In addition to these case-specific instructions, we provided the model with the identified pun type and the list of candidate French homonyms identified by our phonetic-semantic embeddings inference. Our code and full prompts are publicly available at \url{https://github.com/dsgt-arc/joker-2025}.

\subsection{Part 3: Multi-Agent Evaluation of Generated Puns}

\begin{figure}[ht]
    \centerline{\includegraphics[width=\columnwidth]{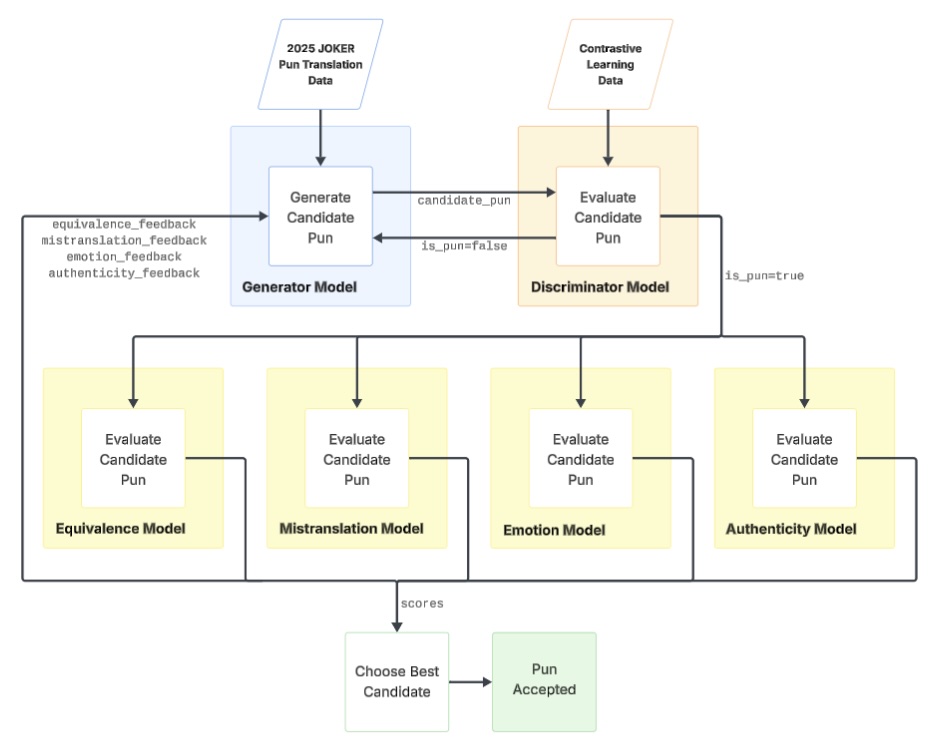}}
    \caption{Multi-agent evaluation of generated puns design.}
    \label{fig-multiagent}
\end{figure}

Assigning roles to prompts can enhance the reasoning capabilities of large language models \cite{shanahan2023roleplaylargelanguagemodels}. Building on this insight, we assign an evaluator role to four distinct agents, each responsible for assessing one of four translation properties, similar to those defined by the Metaphorical Machine Translation Evaluation (MMTE) framework \cite{wang2024mmte}. We adapted MMTE's four categories (quality, metaphorical equivalence, emotion, and authenticity) to better suit our objective of translating humorous language between languages. Each evaluator is explicitly instructed to be fluent in both English and French, with a deep understanding of humor in both languages.

\subsubsection{Equivalence evaluator}

The equivalence evaluator's task is to determine whether the meanings of the source text are maintained in the generated text, and whether the generated pun is humorous. The equivalence evaluator then assigns a rating from 0-2 where:

\begin{itemize}
    \item[2.] \textit{Full equivalence} Both the literal and contextual meanings of the pun remain the same in the translation. The humor, wordplay, and intended effect are fully preserved.
    \item[1.] \textit{Part equivalence} The contextual meaning of the pun is similar in both languages, but the literal meaning of the target word differs. While the translation remains metaphorical, the wordplay may be altered.
    \item[0.] \textit{Non-equivalence} The contextual meaning is somewhat preserved, but the translation is no longer metaphorical. The literal meaning of the target words differs significantly, resulting in a loss of the original wordplay.
\end{itemize}

\subsubsection{Mistranslation evaluator}

The mistranslation evaluator's task is to assess whether the literal meaning of the pun's wordplay is similar in both languages, but the contextual meaning or intended humor is lost or altered in translation. The mistranslation evaluator then assigns a rating from 0-2 where:

\begin{itemize}
    \item[2.] Both the literal and contextual meanings are similar in both the source text and the translation
    \item[1.] The literal meaning of the pun’s wordplay is similar in both the source text and the translation, but the translation fails to convey the contextual meaning or intended humor of the original pun.
    \item[0.] The pun’s wordplay is mistranslated, meaning that both the literal and contextual meanings differ between the source and translation, resulting in a complete loss of the intended pun or humor.
\end{itemize}

\subsubsection{Emotion evaluator}

The emotion evaluator's task is to assess to what extent the original pun's wordplay and its translation convey different amounts of emotion. The emotion evaluator then assigns a rating from 0-1 where:

\begin{itemize}
    \item[0.] Less emotion compared to the original 
    \item[0.] More emotion compared to the original
    \item[1.] Same emotion compared to the original
\end{itemize}

\subsubsection{Authenticity evaluator}

The authenticity evaluator's task is to assess to what extent the translated pun reads like standard, well-edited language, such that the pun would be understood by a native speaker of the French language. The authenticity evaluator then assigns a rating from 0-4 where:
\begin{itemize}
    \item[0.] Not at all likely
    \item[1.] Not Very Likely
    \item[2.] Somewhat Likely
    \item[3.] Very Likely
    \item[4.] Extremely Likely
\end{itemize}

The evaluators then entered a refinement loop for the translated puns, providing feedback at each iteration. The initial translation was generated using our contrastive learning method. This iterative process was repeated five times, and the translation with the highest average evaluation score across all criteria was selected as the final output. Our code and full prompts are publicly available at \url{https://github.com/dsgt-arc/joker-2025}.
\section{Results}

This section reports the results of our experiments, including results from intermediate steps in our process, as well as the final results of our submissions to CLEF 2025 JOKER Task 2.

\subsection{Part 1: Baseline Pun Generation with Contrastive Learning}

Our first two submissions to the shared task used baseline pun generation with contrastive learning.

\subsubsection{Contrastive learning pre-training}

We tested the contrastive learning dataset by sampling 25 positive examples, 25 negative examples, and 450 labeled test examples from the data. We performed several-shot prompting with \texttt{o4-mini} using the positive and negative examples. For the 450 labeled test examples, the model identified 225 out of 225 of the negative examples correctly (100\% accuracy) and 223 out of 225 positive examples correctly (99.11\% accuracy).

\subsubsection{Unsupervised cross-lingual LLM pun generation evaluated using contrastive learning}

Table \ref{tab-generate} shows the results of baseline pun generation for the training dataset using five different models. We evaluated them using \texttt{o4-mini} and \texttt{gemini-2.5-flash} with contrastive learning. As expected, the \texttt{o4-mini} evaluator slightly preferred \texttt{o4-mini} generated puns. Meanwhile, \texttt{gemini-2.5-flash} reported significantly higher levels of success for puns generated by \texttt{deepseek-reasoner} and \texttt{claude-sonnet-4}. However, both evaluator models generally agreed on the relative success of each generator model. \texttt{o4-mini} was the clear leader in generating successful puns.

\begin{table*}[t]
  \caption{A comparative analysis of different models evaluated by various evaluation models. The data shows performance scores, where higher values indicate better performance.}
  \label{tab-generate}
  \begin{tabular}{l c c c}
    \toprule
    & \multicolumn{2}{c}{\textbf{Evaluation Model}} & \textbf{Re-generated} \\
    \cmidrule(lr){2-3} \cmidrule(lr){4-4}
    \textbf{Model} & \textbf{o4-mini} & \textbf{gemini-2.5-flash} & \textbf{gemini-2.5-flash} \\
    \midrule
    o4-mini & 0.9902 & 0.9800 & 1.0000 \\
    gemini-2.5-pro & 0.9275 & 0.9379 & \\
    claude-sonnet-4 & 0.8087 & 0.8575 & \\
    deepseek-reasoner & 0.7917 & 0.8583 & \\
    mistral-medium-2505 & 0.7801 & 0.7644 & 0.9896 \\
    \bottomrule
  \end{tabular}
\end{table*}

For two generator models, we processed the full training and test dataset, using a loop to retry generation up to 10 times until the \texttt{gemini-2.5-flash} evaluator agreed that the generated text contained a pun. We chose \texttt{o4-mini} because it produced the best results on the training dataset, and \texttt{mistral-medium-2505} because of its speed and because it is a natively French model. \texttt{o4-mini} achieved a 100\% success rate with up to 10 retries, while \texttt{mistral-medium-2505} had a few failures even after 10 retries, resulting in a 98.96\% success rate.

\subsubsection{Part 1 shared task submissions}

As expected, the results of the automated evaluation (BLEU and BERTScore) of our baseline shared task submissions were quite low. We deliberately chose to prioritize preservation wordplay and idiomatic translation over literally representing each of the words in the source text. 

Interestingly, while our contrastive learning evaluation ranked \texttt{o4-mini} higher based on containing wordplay, \texttt{mistral-medium-2505} ranks higher based on similarity to the source text. This result matches our observation that \texttt{o4-mini} tended to produce completely different puns for several inputs. The results for the shared task automated metrics are shown in Tables \ref{tab:baseline_bleu} and \ref{tab:baseline_bert}.

\begin{table*}[h!]
\centering
\caption{BLEU automated results for baseline generation with contrastive learning.}
\label{tab:baseline_bleu}
\begin{tabular}{lccccccc}
\toprule
\textbf{Model} & \textbf{Score} & \textbf{1-gram} & \textbf{2-gram} & \textbf{3-gram} & \textbf{4-gram} & \textbf{Ratio} & \textbf{Rank} \\
\midrule
mistral-medium-2505 & 14.94 & 37.47 & 17.74 & 10.90 & 6.88 & 1.2635 & 46/51 \\
o4-mini             & 8.15  & 29.13 & 9.80  & 5.17  & 2.99  & 1.0822 & 47/51 \\
\bottomrule
\end{tabular}
\end{table*}

\begin{table*}[h!]
\centering
\caption{BERTScore automated results for baseline generation with contrastive learning.}
\label{tab:baseline_bert}
\begin{tabular}{lcccc}
\toprule
\textbf{Model} & \textbf{Precision} & \textbf{Recall} & \textbf{F1-Score} & \textbf{Rank} \\
\midrule
mistral-medium-2505 & 0.776  & 0.7912 & 0.783  & 46/51 \\
o4-mini             & 0.7382 & 0.7395 & 0.7385 & 49/51 \\
\bottomrule
\end{tabular}
\end{table*}

The results of human evaluation for our baseline submissions were also low. We expected that the creative capabilities of out-of-the-box LLMs would perform better than other solutions, but the results indicate that creative freedom without the constraints provided by our more complex solutions resulted in puns that were often not closely related to the original. The results for the shared task human evaluation are shown in Table \ref{tab:baseline_human}.

\begin{table*}[h!]
\centering
\caption{Human evaluation results for baseline generation with contrastive learning.}
\label{tab:baseline_human}
\begin{tabular}{lcccc}
\toprule
\textbf{Model} & \textbf{Count} & \textbf{Location} & \textbf{Percent} & \textbf{Rank} \\
\midrule
mistral-medium-2505 & 1682 & 60 & 3.57 & 45/51 \\
o4-mini             & 1682 & 18 & 1.07 & 49/51 \\
\bottomrule
\end{tabular}
\end{table*}

\subsection{Part 2: Guided Chain-of-Thought with Phonetic-Semantic Embeddings}

Our second submission to the shared task used guided chain-of-thought with phonetic-semantic embeddings.

\subsubsection{Identification of pun word and pun type}
The first step in annotating the data in order to guide chain-of-thought prompting was to accurately identify the location of the pun word as well as the type of pun (homographic or homophonic). Table \ref{tab:identify} reports the similarity scores between the identified pun locations and pun types from several different models and our human annotated dataset. 

The latest reasoning models—\texttt{gemini-2.5-pro} and \texttt{o3}—clearly excel at this task. It is notable that the pun location scores are consistently higher for all models than the pun type scores. This is because our binary classification of pun types is too simplistic. During the human annotation process, we found many ambiguous cases where an implied word may be from the same root, and could take the same grammatical form, but may seem more natural in another grammatical form. It was difficult in such cases to draw a firm line between homographic and homophonic puns. And those cases were frequently where the models disagreed with our decisions in various directions.

Overall, these results show that LLMs have become extremely proficient at the pun location task.

\begin{table*}[h!]
\centering
\caption{Identification of pun word and pun type versus manual annotations by model for the training dataset.}
\label{tab:identify}
\begin{tabular}{lcccccccc}
\toprule
& \multicolumn{4}{c}{\textbf{Pun Location}} & \multicolumn{4}{c}{\textbf{Pun Type}} \\
\cmidrule(lr){2-5} \cmidrule(lr){6-9}
\textbf{Model} & \textbf{Accuracy} & \textbf{Precision} & \textbf{Recall} & \textbf{F1-Score} & \textbf{Accuracy} & \textbf{Precision} & \textbf{Recall} & \textbf{F1-Score} \\
\midrule
gemini-2.5-pro    & 0.9687 & 0.9974 & 0.9687 & 0.9685 & 0.8882 & 0.9007 & 0.8882 & 0.8868 \\
o3                & 0.9473 & 0.9949 & 0.9473 & 0.9463 & 0.8811 & 0.8999 & 0.8811 & 0.8790 \\
o4-mini           & 0.9330 & 0.9915 & 0.9330 & 0.9319 & 0.8811 & 0.8929 & 0.8811 & 0.8797 \\
claude-sonnet-4   & 0.9274 & 0.9895 & 0.9274 & 0.9249 & 0.8590 & 0.8812 & 0.8590 & 0.8588 \\
gemini-2.5-flash  & 0.8868 & 0.9897 & 0.8868 & 0.8875 & 0.8675 & 0.8990 & 0.8675 & 0.8748 \\
gpt-4.1           & 0.8817 & 0.9903 & 0.8817 & 0.8782 & 0.8217 & 0.8649 & 0.8217 & 0.8217 \\
\bottomrule
\end{tabular}
\end{table*}

\subsubsection{Translation of pun word and meanings}
The next step in the annotation process was to translate the identified pun word and its meanings, in the form of synonyms lists, into French. We evaluated multiple models on this task using bilingual English/French embeddings and calculating the cosine similarity between the source and target words. Table \ref{tab:translate} shows the results.

\begin{table*}[h!]
\centering
\caption{Translation of pun word and meanings versus bi-lingual embeddings cosine similarities for the training dataset. Cosine Similarity refers to the mean of cosine similarities for all words translated by each model. Top Quartile identifies top 25\% of cosine similarities, and Problems identifies the percentage of words that were not properly translated by that model.}
\label{tab:translate}
\begin{tabular}{lcccc}
\toprule
\textbf{Model} & \textbf{Cosine Similarity} & \textbf{Variance} & \textbf{Top Quartile} & \textbf{Problems} \\
\midrule
gemini-2.5-pro       & 0.8178 & 0.0098 & 0.725  & 0.01   \\
o4-mini              & 0.8128 & 0.0113 & 0.7367 & 0.0    \\
claude-sonnet-4      & 0.8073 & 0.0133 & 0.7031 & 0.0069 \\
mistral-medium-2505  & 0.7926 & 0.0145 & 0.6636 & 0.0243 \\
gemini-2.5-flash     & 0.7856 & 0.0165 & 0.6667 & 0.1767 \\
google-translate     & 0.7751 & 0.0092 & 0.61   & 0.0    \\
\bottomrule
\end{tabular}
\end{table*}

Because the results for \texttt{gemini-2.5-pro} and \texttt{o4-mini} were so close, we decided to proceed with \texttt{o4-mini} for translating the full test dataset, due to its lower cost, faster speed, and lower propensity for errors. Notably, the only machine translation model we tested, \texttt{google-translate}, performed worse than the other models.

\subsubsection{Part 2 shared task submission}

The results for the shared task automated metrics are shown in Tables \ref{tab:cot_bleu} and \ref{tab:cot_bert}. Again these are quite low. We deliberately chose to prioritize preservation wordplay and idiomatic translation over literally representing each of the words in the source text.

This submission improved upon the baseline submissions, likely because the constraints we placed on the generated translations forced them to follow the source text more closely in many cases. Specifically, we guided the model to use words with meanings that were close to the original. 

\begin{table*}[h!]
\centering
\caption{BLEU automated results for guided chain-of-thought with phonetic-semantic embeddings.}
\label{tab:cot_bleu}
\begin{tabular}{ccccccc}
\toprule
\textbf{Score} & \textbf{1-gram} & \textbf{2-gram} & \textbf{3-gram} & \textbf{4-gram} & \textbf{Ratio} & \textbf{Rank} \\
\midrule
16.52 & 39.85 & 19.78 & 12.12 & 7.79 & 1.0645 & 45/51 \\
\bottomrule
\end{tabular}
\end{table*}

\begin{table*}[h!]
\centering
\caption{BERTScore automated results for guided chain-of-thought with phonetic-semantic embeddings.}
\label{tab:cot_bert}
\begin{tabular}{cccc}
\toprule
\textbf{Precision} & \textbf{Recall} & \textbf{F1-Score} & \textbf{Rank} \\
\midrule
0.778 & 0.7915 & 0.7842 & 45/51 \\
\bottomrule
\end{tabular}
\end{table*}

However, our guided chain-of-thought with phonetic-semantic embeddings submission outperformed all other teams in the shared task and was only beaten by our final submission (discussed below). This run placed near the bottom of the rankings when evaluated with automated metrics because those metrics measure literalness. But our aim was primarily to produce humorous, non-literal translations, and that approach was recognized by the human evaluators.

\begin{table*}[h!]
\centering
\caption{Human evaluation results for guided chain-of-thought with phonetic-semantic embeddings.}
\label{tab:cot_human}
\begin{tabular}{cccc}
\toprule
\textbf{Count} & \textbf{Location} & \textbf{Percent} & \textbf{Rank} \\
\midrule
1682 & 132 & 7.85 & 2/51 \\
\bottomrule
\end{tabular}
\end{table*}

\subsection{Part 3: Multi-Agent Evaluation of Generated Puns}

Our final submission to the shared task used multi-agent evaluation of generated puns.

\subsubsection{Evaluation scores and number of retries}

We averaged the scores assigned by each of the four evaluator models where the maximum averaged score is 2.25 and the threshold for acceptance is 2.0. After 5 iterations, if the threshold is not met, the current maximum score is accepted.

Table \ref{tab:iterative_process} shows the count for each iteration, the mean score, the most frequent score (mode), and the variance broken down by iteration and totaled.

\begin{table}[h!]
\centering
\caption{Best averaged scores assigned by evaluators (maximum of 2.25) and number of iterations used.}
\label{tab:iterative_process}
\begin{tabular}{lccccc}
\toprule
\textbf{Iteration} & \textbf{Count} & \textbf{Percent} & \textbf{Mean} & \textbf{Mode} & \textbf{Variance} \\
\midrule
\textbf{1}     & 727  & 0.160 & 1.238 & 1.0  & 0.129 \\
\textbf{2}     & 1196 & 0.264 & 1.862 & 2.25 & 0.176 \\
\textbf{3}     & 1043 & 0.230 & 1.789 & 2.0  & 0.160 \\
\textbf{4}     & 827  & 0.182 & 1.742 & 2.0  & 0.160 \\
\textbf{5}     & 744  & 0.164 & 1.737 & 2.0  & 0.161 \\
\midrule
\textbf{Total} & 4537 & 1.000 & 1.703 & 2.25 & 0.202 \\
\bottomrule
\end{tabular}
\end{table}

The maximum score was achieved most often after 2 iterations. Puns that required more than 2 iterations were less likely to achieve the maximum score. The mean score is lower than the threshold because these numbers include runs selected after failing to meet the threshold 5 times. It is clear from the numbers that iteration 1 was the most frequently chosen iteration in that scenario.

\subsubsection{Part 3 shared task submission}

The results for the shared task automated metrics are shown in Tables \ref{tab:multiagent_bleu} and \ref{tab:multiagent_bert}. This submission produced the highest BLEU scores and BERTScores among our submissions. This is because two of the four evaluator models (equivalence and mistranslation) check for similarity to the source text. However, as with our other runs these results for automated metrics are still quite low because they measure literalness and our aim was to preserve the non-literal nature of the puns.

\begin{table*}[h!]
\centering
\caption{BLEU results for multi-agent evaluation of generated puns.}
\label{tab:multiagent_bleu}
\begin{tabular}{ccccccc}
\toprule
\textbf{Score} & \textbf{1-gram} & \textbf{2-gram} & \textbf{3-gram} & \textbf{4-gram} & \textbf{Ratio} & \textbf{Rank} \\
\midrule
21.41 & 46.61 & 25.26 & 16.35 & 10.92 & 1.0375 & 41/51 \\
\bottomrule
\end{tabular}
\end{table*}

\begin{table*}[h!]
\centering
\caption{BERTScore results for multi-agent evaluation of generated puns.}
\label{tab:multiagent_bert}
\begin{tabular}{cccc}
\toprule
\textbf{Precision} & \textbf{Recall} & \textbf{F1-Score} & \textbf{Rank} \\
\midrule
0.7984 & 0.8157 & 0.8066 & 41/51 \\
\bottomrule
\end{tabular}
\end{table*}

Our multi-agent evaluation of generated puns submission outperformed all other submissions in the shared task by a relatively substantial margin when evaluated by humans. 

\begin{table*}[h!]
\centering
\caption{Human evaluation results for multi-agent evaluation of generated puns.}
\label{tab:multiagent_human}
\begin{tabular}{cccc}
\toprule
\textbf{Count} & \textbf{Location} & \textbf{Percent} & \textbf{Rank} \\
\midrule
1682 & 156 & 9.27 & 1/51 \\
\bottomrule
\end{tabular}
\end{table*}
\section{Discussion}

We designed our experiments with the aim of achieving high human evaluation scores at the expense of high automated evaluation scores. This was a calculated decision based on a theory of translation that prioritizes conveying a sense of the linguistic creativity in the source text over literally translating each word. For that reason, our scores in the automated evaluation were near the bottom of the among submissions to the shared task. Meanwhile, we earned the top two spots in the shared task competition using human evaluation metrics.

\subsection{Observations about the shared task results}

Our top ranking results clearly show the need for emphasis on preserving the humor and non-literal nature of the source text when translating puns. Many other teams fine-tuned language models on the JOKER training data and achieved much higher BLEU and BERTScores than we did. Our approach was largely unsupervised and made relatively little use of the French training data in the JOKER dataset, yet still outperformed the supervised approaches because of our emphasis on translation theory.

The next most interesting result is that our baseline generation models performed poorly (near the bottom of the human evaluation rankings). Because our improved models built directly on the o4-mini baseline model, this result empirically demonstrates the value of both our chain-of-thought with phonetic-semantic embeddings and multi-agent evaluation of generated puns approaches. This result also shows that it was not simply the improvements LLMs have made in recent years that enabled our models to perform so well.

However, it remains to be seen how well the baseline models may have performed when evaluated with more nuanced human evaluation metrics. The metrics used in the JOKER shared task focused primarily on whether the pun word was translated with an appropriate homonymic gloss. Our improved models constrained the outputs to match the original text more closely than our baseline models did. It is possible that the baseline models produced humorous and appropriate translations in many cases, but were not close enough to the original to be considered positive results.

We found it interesting that our multi-agent evaluation of generated puns submission outperformed our guided chain-of-thought with phonetic-semantic embeddings submission. This result shows that LLMs are becoming more and more capable as evaluators. This result also shows promise for the future of evaluating machine-translated puns. Current automated evaluation metrics are sorely lacking, and we believe our approach could be further improved to become a state-of-the-art pun translation evaluation metric.

\subsection{Observations about our experiments}

Some other notable observations about our experiments include the following: 

Large language models have become very good at the task of locating and explaining puns. The latest reasoning models achieved nearly 100\% accuracy for the task of locating the pun word. And while we did not quantify the accuracy of the generated synonyms lists representing each of the two meanings of the pun word, we observed very high degrees of correlation between the identified meanings and those we manually identified.

Feedback loops were quite helpful in multiple prompting tasks. When discriminating between puns and non-puns with the contrastive learning dataset, it brought the percentage of texts identified as containing a pun to nearly 100\%. Creating more nuanced feedback loops improved upon our initial automated results, and we expect that it will improve upon our human-annotated results in previous submissions as well.

Some models are more prone to free-flowing creativity than others. Our baseline \texttt{o4-mini} run often produced puns entirely unrelated to the original. While this trends in the direction we were aiming for (less literal translations), it sometimes went further than we anticipated.

Identifying homonyms using phonetic-semantic embeddings produced viable French pun word candidates for many pun sentences. But in the majority of cases no candidate was found that met our thresholds. This confirms the long-standing observation homonyms often do not have cross-lingual counterparts. Our approach would benefit from pursuing further levels of Low's polygonal algorithm \cite{low2011translating}.

Frontier LLMs outperformed even Google Translate on the task of translating lists of words from English to French. Because these translations were meant to be literal translations, we expected Google Translate to excel and were surprised by this result.

Finally, puns come in many forms and too simplistic a classification can hinder efforts at translating them.

\section{Future Work}

We believe that combining our approaches from Parts 2 and 3 would produce results that would outperform both of them. In the present experiments, we used the baseline \texttt{o4-mini} runs as inputs to Part 3, and we evaluated Part 2 only using the simple contrastive learning discriminator. In the future, we would like to implement our guided chain-of-thought pipeline together with a more robust evaluator.

Our results in Part 2 demonstrated the need to implement additional levels of Low's polygonal algorithm. Our implementation extended to the pentagon level, but we are confident that better homonyms could be found with additional levels.

Our binary classification of pun types (homophonic and homographic) was a limiting factor for our Part 2 process. Implementing more nuanced categories and generation prompts based on those categories, should produce improved results.

Our implementation did not account for the presence of multiple puns in a given pun sentence. We observed several cases where this occurred in the dataset, but limited our approach to only one pun word per input. Future work should account for the many varied linguistic forms that puns may take.

While our results outperformed other teams' submissions to the JOKER shared task, there remains significant room for improvement. These suggestions are likely to lead to further advances in future work.
\section{Conclusions}

This study presents a multi-stage methodology for translating puns by prioritizing the preservation of humor and wordplay over literal equivalence. Our approach progresses from baseline models to a guided chain-of-thought pipeline using phonetic-semantic embeddings and finally to a multi-agent refinement loop. We demonstrated that large language models are adept at nuanced linguistic sub-tasks, and that focused improvements to both pun generation and evaluation yield promising results. Ultimately, we showed that shifting the goal from literal accuracy to functional equivalence is crucial for successful machine translation of wordplay.

\begin{acknowledgments}
We thank the Data Science at Georgia Tech (DS@GT) CLEF competition group for their support.
This research was supported in part through research cyberinfrastructure resources and services provided by the Partnership for an Advanced Computing Environment (PACE) at the Georgia Institute of Technology, Atlanta, Georgia, USA \cite{PACE}. 
\end{acknowledgments}

\section*{Declaration on Generative AI}
 During the preparation of this work, the authors used \texttt{gemini-2.5-pro} in order to generate LaTeX equations and to check grammar and spelling. After using this tool, the authors reviewed and edited the content as needed and take full responsibility for the publication’s content. 

\bibliography{main}
\end{document}